\documentclass[10pt,twocolumn,letterpaper]{article}

\usepackage{cvpr}
\usepackage{times}
\usepackage{epsfig}
\usepackage{graphicx}
\usepackage{amsmath}
\usepackage{amssymb}

\usepackage{subfig} 
\usepackage{algorithm}
\usepackage{algorithmic}
\usepackage{threeparttable}
\usepackage{booktabs}
\usepackage{multirow}
\usepackage{comment}
\usepackage{tikz}
\usetikzlibrary{bayesnet}

\usepackage[font={small,it}]{caption}

\newcommand\numberthis{\addtocounter{equation}{1}\tag{\theequation}}

\usepackage[pagebackref=true,breaklinks=true,letterpaper=true,colorlinks,bookmarks=false]{hyperref}

\makeatletter
\newcommand{\smallsym}[2]{#1{\mathpalette\make@small@sym{#2}}}
\newcommand{\make@small@sym}[2]{%
  \vcenter{\hbox{$\m@th\downgrade@style#1#2$}}%
}
\newcommand{\downgrade@style}[1]{%
  \ifx#1\displaystyle\scriptstyle\else
    \ifx#1\textstyle\scriptstyle\else
      \scriptscriptstyle
  \fi\fi
}
\makeatother

\newcommand{\bm}[1]{{\mathbf{#1}}}
\newcommand{\bx}{{\mathbf{x}}}
\newcommand{\by}{{\mathbf{y}}}
\newcommand{\bz}{{\mathbf{z}}}
\newcommand{\bu}{{\mathbf{u}}}

\def\ie{\emph{i.e.~}}
\def\eg{\emph{e.g.~}}
\def\etc{\emph{etc}}
\def\etal{\emph{et al.}}

\cvprfinalcopy

\begin{document}

\title{Disentangling Latent Hands for Image Synthesis and Pose Estimation} 

\author{Linlin Yang\\
University of Bonn, Germany\\
{\tt\small yangl@cs.uni-bonn.de}
\and Angela Yao\\
National University of Singapore, Singapore\\
{\tt\small ayao@comp.nus.edu.sg}
}

\maketitle
\thispagestyle{empty}

\begin{abstract}
Hand image synthesis and pose estimation from RGB images are both highly challenging tasks due to the large discrepancy between factors of variation ranging from image background content to camera viewpoint.  To better analyze these factors of variation, we propose the use of disentangled representations and a disentangled variational autoencoder (dVAE) that allows for specific sampling and inference of these factors.  The derived objective from the variational lower bound as well as the proposed training strategy are highly flexible, allowing us to handle cross-modal encoders and decoders as well as semi-supervised learning scenarios.  Experiments show that our dVAE can synthesize highly realistic images of the hand specifiable by both pose and image background content and also estimate 3D hand poses from RGB images with accuracy competitive with state-of-the-art on two public benchmarks.
\end{abstract}

\section{Introduction}
Vision-based hand pose estimation has progressed very rapidly in the past years~\cite{review15,yuan20173d}, driven in part by its potential for use in human-computer interaction applications.  Advancements are largely due to the widespread availability of commodity depth sensors as well as the strong learning capabilities of deep neural networks. As a result, the majority of state-of-the-art methods apply deep learning methods to depth images~\cite{ge2018hand,Ge_multiview,ge20173d,guo2017region,madadi2017occlusion,oberweger2017deepprior++,Oberweger_15deep,wan2017crossing,wan2018dense}. Estimating 3D hand pose from single RGB images, however, is a less-studied and more difficult problem which has only recently gained some attention~\cite{caiweakly,mueller2018ganerated,panteleris2017using,spurr2018cvpr,zimmermann2017learning}.

Unlike depth, which is a 2.5D source of information, RGB inputs have significantly more ambiguities.  These ambiguities arise from the 3D to 2D projection and diverse backgrounds which are otherwise less pronounced in depth images. 
As such, methods which tackle the problem of monocular RGB hand pose estimation rely on learning from large datasets~\cite{zimmermann2017learning}. However, given the difficulties of accurately labelling hand poses in 3D, large-scale RGB datasets collected to date are synthesized~\cite{mueller2018ganerated,zimmermann2017learning}.  Real recorded datasets are much smaller, with only tens of sequences~\cite{tzionas2016ijcv,zhang20163d}.  This presents significant challenges when it comes to learning and motivates the need for strong kinematic and or image priors.  


\begin{figure}[t]
	\centering	
	\includegraphics[width=\linewidth]{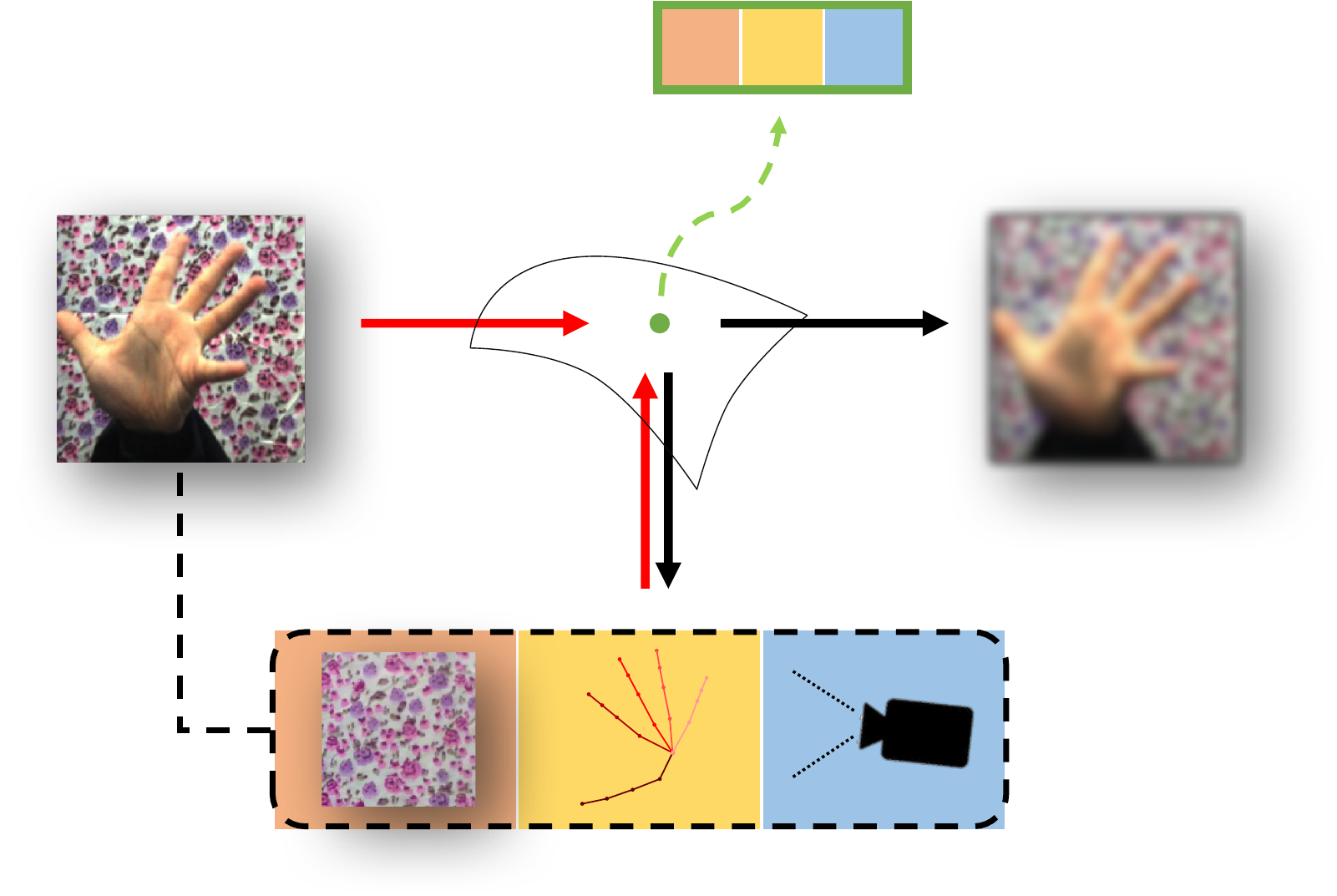}
	\caption{{Illustration of dVAE. The red lines denote variational approximations while the black lines denote the generative model. With the help of labelled factors of variations (\eg pose, viewpoint and image content), we learn a disentangled and specifiable representation for RGB hand images in a VAE framework.}
	}
	\label{fig:pipeline}
	\vspace{-0.5cm}
\end{figure}


Even though straight-forward discriminative approaches have shown great success in accurately estimating hand poses, there has also been growing interest in the use of deep generative models such as adversarial networks (GANs)~\cite{mueller2018ganerated,wan2017crossing} and variational autoencoders (VAEs)~\cite{spurr2018cvpr}.  Generative models can approximate and sample from the underlying distribution of hand poses as well as the associated images, and depending on the model formulation, may enable semi-supervised learning. This is particularly appealing for hand pose estimation, for which data with accurate ground truth can be difficult to obtain.  One caveat, however, is that in their standard formulation, GANs and VAEs learn only black-box latent representations.  Such representations offer little control for conditioning upon human-interpretable factors.  Of the deep generative works presented to date~\cite{mueller2018ganerated,spurr2018cvpr,wan2017crossing}, the latent representations are specifiable only by hand pose.  Consequently it is possible to sample only a single (average) image per pose. 




A recent work combining VAEs and GANs~\cite{de2018dgpose} introduced a conditional dependency structure to learn image backgrounds and demonstrated the possibility of transferring body poses onto different images. Inspired by this work, we would like to learn a similar latent representation that can disentangle the different factors that influence how hands may appear visually, \ie normalized hand pose, camera viewpoint, scene context and background, \etc. 
At the same time, we want to ensure that the disentangled representation remains sufficiently discriminative to make highly accurate estimates of 3D hand pose. 


We present in this paper a disentangled variational autoencoder (dVAE) -- a novel framework for learning disentangled representations of hand poses and hand images. As the factors that we would like to disentangle belong to different modalities, we begin with a cross-modal VAE~\cite{pandey2017variational,spurr2018cvpr} as the baseline upon which we define our dVAE.  By construction, our latent space is a disentangled one, composed of sub-spaces calculated by factors and a training strategy to fuse different latent space into one disentangled latent space. We show how these disentangled factors can be learned from both independent and confounding label inputs.  To the best of our knowledge, our proposed model is the first disentangled representation that is able to both synthesize hand images and estimate hand poses with explicit control over the latent space. A schematic illustration of our dVAE and the disentangled factors is shown in Fig.~\ref{fig:pipeline}.  We summarize our contributions below:





\begin{itemize}
    \item We propose a novel disentangled VAE model crossing different modalities; this model is the first VAE-based model that uses independent factors of variations to learn disentangled representations.
    \item Our dVAE model is highly flexible and handles multiple tasks including RGB hand image synthesis, pose transfer and 3D pose estimation from RGB images.
    \item We enable explicit control over different factors of variation and introduce the first model with multiple degrees of freedom for synthesizing hand images.
    \item We decouple the learning of disentangling factors and the embedding of image content and introduce two variants of learning algorithms for both independent and confounding labels.
\end{itemize}



\section{Related Works}
\subsection{Hand Pose Estimation}
Much of the progress made in hand pose estimation have focused on using depth image inputs~\cite{ge2018hand,Ge_multiview,ge20173d,guo2017region,iqbal2018hand,madadi2017occlusion,moon2018v2v,oberweger2017deepprior++,Oberweger_15deep,wan2017crossing,wan2018dense,wohlke2018model}.  State-of-the-art methods use a convolutional neural network\,(CNN) architecture, with the majority of works treating the depth input as 2D pixels, though a few more recent approaches treat depth inputs as a set of 3D points and or voxels~\cite{ge20173d,ge2018hand,moon2018v2v}. 

Estimating hand poses from monocular RGB inputs is more challenging.  Early methods could recognize only a restricted set of poses~\cite{athitsos2003estimating,wu2000view} or used simplified hand representations instead of full 3D skeletons~\cite{stenger2001model,wu2001capturing}. In more recent approaches, the use of deep learning and CNNs has become common-place~\cite{caiweakly,panteleris2017using,zimmermann2017learning}. 
In~\cite{mueller2018ganerated,spurr2018cvpr}, deep generative models such as variational auto-encoders (VAE)~\cite{spurr2018cvpr} and generative adversarial networks (GANs)~\cite{mueller2018ganerated} are applied, which makes feasible not only to estimate pose, but also generate RGB images from given hand poses.  



Two hand pose estimation approaches~\cite{wan2017crossing,spurr2018cvpr} stand out for being similar to ours in spirit.  They also use shared latent spaces, even though the nature of these spaces are very different.  
Wan~\etal~\cite{wan2017crossing} learns two separate latent spaces, one for hand poses and one for depth images, and uses a one-to-one mapping function to connect the two. 
Spurr~\etal~\cite{spurr2018cvpr} learns a latent space that cross multiple hand modalities, such as RGB to pose and depth to pose. To force the cross-modality pairings onto a single latent space, separate VAEs are learned in an alternating fashion, with one input modality contributing to the loss per iteration. Such a learning strategy is non-ideal, as it 
tends to result in fluctuations in the latent space and has no guarantees for convergence.  Additionally, by assuming all crossing modalities as one-to-one mappings, 
only one image can be synthesized per pose. 



Different from~\cite{wan2017crossing} and~\cite{spurr2018cvpr}, our dVAE learns a single latent space by design. 
We learn the latent space with the different modalities jointly, as opposed to alternating framework of~\cite{spurr2018cvpr}.  
We find that our joint learning is more stable and has better convergence properties.  And because we explicitly model and disentangle image factors, we can handle one-to-many mappings, \ie synthesize multiple images of the same hand pose. 


\subsection{Disentangled Representations}\vspace{-0.1cm}
Disentangled representations separate data according to salient factors of variation and have recently been learned with deep generative models such as VAEs and GANs.  Such representations
have been applied successfully to image editing~\cite{bouchacourt2017multi,de2018dgpose,kulkarni2015deep,narayanaswamy2017learning,shu2017neural,szabo2017challenges}, video generation~\cite{tulyakov2017mocogan} and image-to-image translation~\cite{jha2018disentangling}. 
Several of these works~\cite{shu2017neural,szabo2017challenges,tulyakov2017mocogan,wang2017tag}, however, require specially designed layers and loss functions, making 
the architectures difficult to work with and extend beyond their intended task.


Previous works learning disentangled representations with VAEs~\cite{bouchacourt2017multi,jha2018disentangling,kulkarni2015deep} 
typically require additional weak labels such as grouping information~\cite{bouchacourt2017multi,kulkarni2015deep} and pairwise similarities~\cite{jha2018disentangling}. 
Such labels can be difficult to obtain and are often not defined for continuous variables such as hand pose and viewpoint.  
In \cite{de2018dgpose,narayanaswamy2017learning}, a conditional dependency structure is proposed to train disentangled representations for a semi-supervised learning.  The work of~\cite{de2018dgpose} resembles ours in the sense that they also disentangle pose from appearance; however, their conditional dependency structure is sensitive to the number of factors. As the number of factors grows, the complexity of the network structure increases exponentially.
In comparison to existing VAE  approaches, we are able to learn interpretable and disentangled representations by the shared latent space produced by image and its corresponding factors without additional weak labels.

\section{Methodology}

\subsection{Cross Modal VAE}~\label{sec:cvae}
Before we present how a disentangled latent space can be incorporated into a VAE framework across different modalities, we first describe the original cross modal VAE~\cite{pandey2017variational,spurr2018cvpr}. 
As the name suggests, the cross modal VAE aims to learn a VAE model across two different modalities $\mathbf{x}$ and $\mathbf{y}$. We begin by defining the log probability of the joint distribution $p(\mathbf{x},\mathbf{y})$. Since working with this distribution is intractable, one maximizes the evidence lower bound (ELBO) instead via a latent variable $\mathbf{z}$. Note that $\mathbf{x}$ and $\mathbf{y}$ are assumed to be conditionally independent given the latent $\mathbf{z}$, \ie $\left(\bx\!\perp\!\by\,|\,\bz\right) $. 
\begin{align*}
\log p(\mathbf{x},\mathbf{y}) & \geq \text{ELBO}_{\text{cVAE}}(\mathbf{x},\mathbf{y},\theta_{\bx},\theta_{\by},\phi) \numberthis  \label{eq:crossvae} \\
  & = 
E_{\bz\thicksim q_\phi}\log p_{\theta_\bx}(\mathbf{x}|\mathbf{z}) 
+  E_{\bz\thicksim q_\phi}\log p_{\theta_\by}(\mathbf{y}|\mathbf{z}) \\
& - D_{KL}(q_\phi(\mathbf{z}|\mathbf{x})||p(\mathbf{z})). 
\end{align*}


\noindent Here, $ D_{KL}(\cdot)$ is the Kullback-Leibler divergence.  The variational approximation $q_\phi(\mathbf{z}|\mathbf{x})$ can be thought of as an encoder from $\bx$ to $\bz$, while $p_{\theta_\bx}(\mathbf{x}|\mathbf{z})$ and $p_{\theta_\by}(\mathbf{y}|\mathbf{z})$ can be thought of as decoders from $\bz$ to $\bx$ and $\bz$ to $\by$ respectively.   $p(\mathbf{z}) = \mathcal{N}(\mathbf{0},\mathbf{I})$ is a Gaussian prior on the latent space. 

In the context of hand pose estimation, $\bx$ would represent the RGB or depth image modality and $\by$ the hand skeleton modality.  One can then estimate hand poses from images by encoding the image $\bx$ into the latent space and decoding the corresponding 3D hand pose $\mathbf{y}$.  A variant of this model was applied in~\cite{spurr2018cvpr} and shown to successfully estimate hand poses from RGB images or depth images.

\subsection{Disentangled VAE}~\label{sec:dvae}

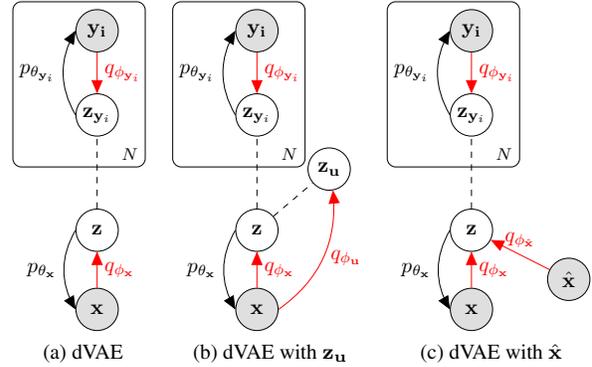
\begin{figure}[t!]
\centering
\subfloat[dVAE]{\label{fig:dvae_simp}
    \scalebox{0.8}{\begin{tikzpicture}[baseline=(X.base),x=1.3cm,y=1.6cm] 
	\node[obs]                 (X)       {$\bm{x}$} ; 
	\node[latent, above=of X, yshift=-1.0cm]  (z)       {$\bm{z}$} ; 	
	\node[obs, above=of z, yshift=1.0cm]     (y)   {$\bm{y_i}$} ; 	
	\node[latent, above=of z,yshift=-0.4cm]   (zy)       {$\bm{z}_{\bm{y}_i}$} ;  
	\node[invis, right=of y, xshift=-1.7cm]  (i1)      {} ;	
	\node[invis, left=of y, xshift=1.1cm]  (i2)      {} ;	
    \plate {plate1} { (y) (zy) (i1) (i2) } {$N$};
	
	\draw[-,dashed](z)--(zy) ; 
	
	\draw[->,>={triangle 45},red] (X) to node[right]{$q_{\phi_\bm{x}}$} (z) ;
	\draw[->,>={triangle 45},red] (y) to node[right]{$q_{\phi_{\bm{y}_i}}$} (zy) ;
	\draw[->,>={triangle 45}] (zy.west) to[bend left] node[left]{$p_{\theta_{\bm{y}_i}}$} (y.west) ;
	\draw[->,>={triangle 45}] (z.west) to[bend right] node[left]{$p_{\theta_\bm{x}}$} (X.west) ;	
\end{tikzpicture}}
    }
\subfloat[dVAE with $\bz_{\bu}$ ]{\label{fig:dvae_gen}
   \scalebox{0.8}{ \begin{tikzpicture}[baseline=(X.base),x=1.3cm,y=1.6cm] 
	\node[obs]                 (X)       {$\bm{x}$} ; 
	\node[latent, above=of X, yshift=-1.0cm]  (z)       {$\bm{z}$} ; 	
	\node[obs, above=of z, yshift=1.0cm]     (y)   {$\bm{y_i}$} ; 	
	\node[latent, above=of z,yshift=-0.4cm]   (zy)       {$\bm{z}_{\bm{y}_i}$} ;  
	\node[latent, above=of z, xshift=1.2cm,yshift=-1.3cm]  (zu)      {$\bm{z_u}$} ;
	\node[invis, right=of y, xshift=-1.7cm]  (i1)      {} ;	
	\node[invis, left=of y, xshift=1.1cm]  (i2)      {} ;	
    \plate {plate1} { (y) (zy) (i1) (i2) } {$N$};
	
	\draw[-,dashed](z)--(zy) ; 
	\draw[-,dashed](z)--(zu) ;
	
	\draw[->,>={triangle 45},red] (X) to node[right]{$q_{\phi_\bm{x}}$} (z) ;
	\draw[->,>={triangle 45},red] (X) (X.east) to [bend right]  node [right] {$q_{\phi_\bm{u}}$} (zu) ;	
	\draw[->,>={triangle 45},red] (y) to node[right]{$q_{\phi_{\bm{y}_i}}$} (zy) ;
	\draw[->,>={triangle 45}] (zy.west) to[bend left] node[left]{$p_{\theta_{\bm{y}_i}}$} (y.west) ;
	\draw[->,>={triangle 45}] (z.west) to[bend right] node[left]{$p_{\theta_\bm{x}}$} (X.west) ;	
\end{tikzpicture} }
   }
\subfloat[dVAE with $\hat{\bx}$]{\label{fig:dvae_simp2}
    \scalebox{0.8}{\begin{tikzpicture}[baseline=(X.base),x=1.3cm,y=1.6cm] 
\node[obs]                 (X)       {$\bm{x}$} ; 
\node[obs, right=of X, xshift=-0.4cm, yshift=0.5cm]  (xhat)      {$\hat{\bm{x}}$} ;  
\node[latent, above=of X, yshift=-1.0cm]  (z)       {$\bm{z}$} ; 	
\node[obs, above=of z, yshift=1.0cm]     (y)   {$\bm{y_i}$} ; 	
\node[latent, above=of z,yshift=-0.4cm]   (zy)       {$\bm{z}_{\bm{y}_i}$} ;  

\node[invis, right=of y, xshift=-1.7cm]  (i1)      {} ;	
\node[invis, left=of y, xshift=1.1cm]  (i2)      {} ;	
\plate {plate1} { (y) (zy) (i1) (i2) } {$N$};

\draw[-,dashed](z)--(zy) ; 

\draw[->,>={triangle 45},red] (X) to node[right]{$q_{\phi_\bm{x}}$} (z) ;
\draw[->,>={triangle 45},red] (xhat) to node[above](phi) {$q_{\phi_{\hat{\bm{x}}}}$} (z) ;
\draw[->,>={triangle 45},red] (y) to node[right]{$q_{\phi_{\bm{y}_i}}$} (zy) ;
\draw[->,>={triangle 45}] (zy.west) to[bend left] node[left]{$p_{\theta_{\bm{y}_i}}$} (y.west) ;

\draw[->,>={triangle 45}] (z.west) to[bend right] node[left]{$p_{\theta_\bm{x}}$} (X.west) ;	

\end{tikzpicture}}
    }
\caption{Graphical models of disentangled VAEs. The shaded nodes represent observed variables while un-shaded nodes are latent. The red and black solid lines denote variational approximations $q_{\phi}$ or encoders, and the generative models $p_{\theta}$ or decoders respectively.  The dashed lines denote deterministically constructed variables. 
Figure best viewed in colour.
\vspace{-0.4cm}
}\label{fig:diagram}
\end{figure}

In our disentangled VAE, we define a latent variable $\bz$ which can be deterministically decomposed into $N+1$ independent factors $\{\bz_{\by_1}, \bz_{\by_2}, ..., \bz_{\by_N}, \bz_{\bu}\}$.  Of these factors, $\{\bz_{\by_i}\}_{i=1...N}$ are directly associated with observed variables $\{\by_i\}_{i=1...N}$.  $\bz_{\bu}$ is an extra latent factor which is not independently associated with any observed variables; it may or may not be included (compare Fig.~\ref{fig:dvae_simp} versus Fig.~\ref{fig:dvae_gen}). 

\textbf{Fully specified latent $\bz$:}
We begin first by considering the simplified case in which $\bz$ can be fully specified by $\bz_{\by_i}$ without $\bz_\bu$, \ie all latent factors can be associated with some observed $\by_i$.  For clarity, we limit our explanation to $N\!=\!2$, though the theory generalizes to higher $N$ as well.  Our derivation can be separated into a disentangling step and an embedding step. In the \emph{\textbf{disentangling step}}, we first consider the joint distribution between $\bx$, $\by_1$ and $\by_2$.  The evidence lower bound of this distribution can be defined as:
\begin{align*}
\log p(\bx,& \by_1,\by_2)  \geq 
\text{ELBO}_{\text{dis}}
(\bx,\by_1,\by_2,\phi_{\by_1},\phi_{\by_2},\theta_{\by_1},\theta _{\by_2},\theta _\bx) \\
& =  \lambda_\bx E_{\bz\thicksim q_{\phi_{\by_1},\phi_{\by_2}}
} \log p_{\theta_\bx}(\bx|\bz)  \\
  & + \lambda_{\by_1} E_{\mathbf{\bz_{\by_1}}\thicksim q_{\phi_{\by_1}}
} \log p_{\theta_{\by_1}}(\by_1|\mathbf{\bz_{\by_1}}) \\
  & + \lambda_{\by_2} E_{\mathbf{\bz_{\by_2}} \thicksim q_{\phi_{\by_2}}
} \log p_ {\theta _{\by_2}}(\by_2|\mathbf{\bz_{\by_2}}) \\
 & - \beta D_{KL}\left(q_{\phi_{\by_1},\phi_{\by_2}}(\bz|\by_1,\by_2)||p(\bz)\right),  \numberthis  \label{eq:disentangle}
\end{align*}
where the $\lambda$s and $\beta$ are additional hyperparameters added to trade off between latent space capacity and reconstruction accuracy,  as recommended by the $\beta$ trick~\cite{higgins2016beta}.

The ELBO in Eq.~\ref{eq:disentangle} allows us to define a disentangled $\bz = [\bz_{\by_1}, \bz_{\by_2}]$ based on $\by_1$, $\by_2$ and $\bx$.  In this step, one can learn the encoding and decoding of $\by_i$ to and from $\bz_{\by_i}$, as well as the decoding of $\bz$ to $\bx$.  However, the mapping from $\bx$ to $\bz$ is still missing so we need an additional 
\emph{\textbf{embedding step}} \cite{vedantam2017generative} to learn the encoder 
$q_{\phi_\mathbf{x}}(\mathbf{z}|\mathbf{x})$. 
Keeping all decoders fixed, 
$q_{\phi_\mathbf{x}}(\mathbf{z}|\mathbf{x})$ can be learned by maximizing: 
\begin{align*}
\mathcal{L}(\phi_{\bx} & |\theta_{\by_1},\theta_{\by_2},\theta _{\bx}) 
=-D_{KL}\left(q_{\phi_{\bx}}(\bz|\bx)||p_\theta(\bz|\bx,\by_1,\by_2)
\right)\\
&= \text{ELBO}_{\text{emb}}(\bx,\by_1,\by_2,\phi_{\bx}) - \log p(\bx,\by_1,\by_2). \numberthis  \label{eq:encoder_x}
\end{align*}

\noindent Since the second term is constant with respect to $\phi_{\bx}$ and the $\theta$'s, the objective simplifies to the following evidence lower bound with $\lambda'$ and $\beta'$ as hyperparameters:
\begin{align*}
\text{ELBO}_{\text{emb}}(\bx,& \by_1,\by_2,\phi_{\bx}) =  \lambda'_{\bx} E_{\bz\thicksim q_{\phi_{\bx}}} \log p_{\theta_{\bx}}(\bx|\bz) \\
& + \lambda'_{\by_1} E_{\bz_{\by_1}\thicksim q_{\phi_{\bx}}}\log p_ {\theta _{\by_1}}(\by_1|\bz_{\by_1}) \\
& + \lambda'_{\by_2} E_{\bz_{\by_2} \thicksim q_{\phi_{\bx}}}\log p_ {\theta _{\by_2}}(\by_2|\bz_{\by_2}) \\
 &- \beta' D_{KL}(q_{\phi_{\bx}}(\bz|\bx)||p(\bz)). \numberthis  \label{eq:objective_x}
\end{align*}

\noindent Combining the disentangling and embedding evidence lower bounds, we get the following joint objective:
\begin{align*}
 \mathcal{L} (\phi_{\bx}, & \phi_{\by_1},\phi_{\by_2},\theta_{\bx},\theta_{\by_1},\theta_{\by_2}) = \\ & \text{ELBO}_{\text{dis}}(\bx,\by_1,\by_2,\phi_{\by_1},\phi_{\by_2},\theta _{\bx},\theta_{\by_1},\theta _{\by_2})\\
+ & \text{ELBO}_{\text{emb}}(\bx,\by_1,\by_2,\phi_{\bx}). \numberthis  \label{eq:objective_all}
\end{align*}

The above derivation shows that the encoding of modality $\bx$ can be decoupled from $\by_1$ and $\by_2$ via a disentangled latent space.
We detail the training strategy for the fully specified version of the dVAE in Alg.\;\ref{alg:alg_original}.
\begin{algorithm}[b!]
	\caption{dVAE learning for fully specified $\bz$.}
	\label{alg:alg_original}
	\begin{algorithmic}[1]
		\REQUIRE $\bx,\by_1,\by_2,\lambda_{\bx},\lambda_{\by_1},\lambda_{\by_2},\beta,T_1,T_2$
		\ENSURE $\phi_{\bx},\phi_{\by_1},\phi_{\by_2},\theta_{\bx},\theta_{\by_1},\theta_{\by_2}$
		\STATE Initialize $\phi_{\bx},\phi_{\by_1},\phi_{\by_2},\theta_{\bx},\theta_{\by_1},\theta_{\by_2}$
		
		\FOR{$t_1 = 1 ,\dots, T_1$ epochs}
		\STATE{Encode $\by_1,\by_2$ to $q_{\phi_{\by_1}}(\bz_{\by_1}|\by_1),q_{\phi_{\by_2}}(\bz_{\by_2}|\by_2)$}
		\STATE{Construct $\bz \gets {[\bz_{\by_1},\bz_{\by_2}]}$}
		\STATE{Decode $\bz$ to $p_{\theta_{\bx}}(\bx|\bz),p_{\theta_{\by_1}}(\by_1|\bz_{\by_1}),p_{\theta_{\by_2}}(\by_2|\bz_{\by_2})$}
		\STATE{Update $\phi_{\by_1},\phi_{\by_2},\theta_{\by_1},\theta_{\by_2},\theta_{\bx}$ via gradient ascent of Eq.\;\ref{eq:disentangle}}
		\ENDFOR
		
		\FOR{$t_2 = 1 ,\dots, T_2$ epochs}
		\STATE{Encode $\bx$ to $q_{\phi_{\bx}}(\bz|\bx)$}
		\STATE{Construct $[\bz_{\by_1},\bz_{\by_2}]  \gets \bz$}
		\STATE{Decode $\bz$ to $p_{\theta_{\bx}}(\bx|\bz),p_{\theta_{\by_1}}(\by_1|\bz_{\by_1}),p_{\theta_{\by_2}}(\by_2|\bz_{\by_2})$}
		\STATE{Update $\phi_{\bx}$ via gradient ascent of Eq.\;\ref{eq:objective_x}}
		\ENDFOR
		
	\end{algorithmic}
\end{algorithm}

\paragraph{Additional $\bz_{\bu}$:} When learning a latent variable model, many latent factors may be very difficult to associate independently with an observation (label), \eg the style of handwritten digits, or the background content in an RGB image~\cite{de2018dgpose,kulkarni2015deep,bouchacourt2017multi}.  Nevertheless, we may still want to disentangle such factors from those which can be associated independently. We model these factors in aggregate form via a single latent variable $\bz_{\bu}$ and show how $\bz_{\bu}$ can be disentangled from the other $\bz_{\by_i}$ which are associated with direct observations $\by_i$. 
For clarity of discussion, we limit $N=1$, such that $\bz = [\bz_{\by_1}, \bz_{\bu}]$.  To disentangle $\bz_{\bu}$ from $\bz$, both of which are specified by a confounding $\bx$, we aim to make $\bz_{\bu}$ and $\by_1$ conditionally independent given $\bz_{\by_1}$ 
To achieve this, we try to make  $p(\by_1|\bz_{\by_1},\bz_{\bu})$ approximately equal to $p(\by_1|\bz_{\by_1})$ and update the encoder and the decoder of $\by_1$ by random sampling of $\bz_{\bu}$ 
and minimizing the distance between $p(\by_1|\bz_{\by_1},\bz_{\bu})$ and $p(\by_1|\bz_{\by_1})$. The training strategy for this is detailed in Alg.~\ref{alg:alg_semi}. 
In this case, the joint distribution of $\bx$ and $\by_1$ has the following evidence lower bound in the \emph{\textbf{disentangling step}} with hyperparameters $\lambda''$ and $\beta''$: 
	\begin{align*}
	\log p(& \bx, \by_1) \geq \text{ELBO}^{\bu}_{\text{dis}}(\bx,\by_1,\phi_{\by_1},\phi_{\bu},\theta _{\by_1},\theta_{\bx}) \\
	 = & \lambda''_{\bx}E_{\bz\thicksim q_{\phi_{\by_1},\phi_{\bu}}}\log p_{\theta_{\bx}}(\bx|\bz) \\
		+ & \lambda''_{\by_1}E_{\bz\thicksim q_{\phi_{\by_1},\phi_{\bu}}}\log p_ {\theta _{\by_1}}(\by_1|\bz) \\ - & \beta''D_{KL}(q_{\phi_{\by_1},\phi_{\bu}}(\bz|\by_1,\bx)||p(\bz)). \numberthis \label{eq:encoder_uv_semi}
	\end{align*}
Note that in the above ELBO, $\bz_{\bu}$ is encoded from $\bx$ by $q_{\phi_{\bu}}$ instead of being specified by some observed label $\bu$, as was done previously in~\cite{kulkarni2015deep,bouchacourt2017multi,de2018dgpose}. 
After this modified disentangling step, we can apply the same embedding step in Eq.~\ref{eq:encoder_x} to learn $q_{\phi_\mathbf{x}}(\mathbf{z}|\mathbf{x})$.

\paragraph{Multiple $\bx$ modalities:}
The situation may arise in which we have multiple input modalities which fully specify and share the latent space of $\bz$, \ie not only an $\bx$ but also an additional $\hat{\bx}$ (see Fig.~\ref{fig:dvae_simp2}).  Here, it is possible to 
first consider the joint distribution between $\bx$, $\by_1$ and $\by_2$, and maximize the ELBO in Eq.~\ref{eq:disentangle} for the disentangling step. To link the two modalities of $\bx$ and $\hat{\bx}$ into the same disentangled latent space and embed $\hat{\bx}$, we can use the following: \begin{equation}
 	\label{eq:encoder_y}
 	\begin{aligned}
 		\mathcal{L}(\phi_{\hat{\bx}}&|\theta _{\bx}, \theta _{\by_1},\theta _{\by_2})=-D_{KL}(q_{\phi_{\hat{\bx}}}(\bz|\hat{\bx})||p_\theta(\bz|\bx,\by_1,\by_2))\\
 		&= \text{ELBO}'_{\text{emb}}(\hat{\bx},\bx,\by_1,\by_2,\phi_{\hat{\bx}}) - \log p(\bx,\by_1,\by_2).
 	\end{aligned}
 \end{equation}
Similar to Eq.~\ref{eq:objective_x}, we get the following evidence lower bound with $\lambda'''$ and $\beta'''$ as hyperparameters:
\begin{align*}
\text{ELBO}'_{\text{emb}}(\hat{\bx},\bx&, \by_1,\by_2,\phi_{\hat{\bx}}) =  \lambda'''_{\bx} E_{\bz\thicksim q_{\phi_{\hat{{\bx}}}}} \log p_{\theta_{{\bx}}}(\bx|\bz) \\
& + \lambda'''_{\by_1} E_{\bz_{\by_1}\thicksim q_{\phi_{\hat{\bx}}}} \log p_ {\theta _{\by_1}}(\by_1|\bz_{\by_1}) \\
& + \lambda'''_{\by_2} E_{\bz_{\by_2} \thicksim q_{\phi_{\hat{\bx}}}} \log p_ {\theta _{\by_2}}(\by_2|\bz_{\by_2}) \\
 &- \beta''' D_{KL}(q_{\phi_{\hat{\bx}}}(\bz|\hat{\bx})||p(\bz)). \numberthis  \label{eq:objective_cross}
\end{align*}

\noindent For learning, one simply encodes $\hat{\bx}$ with $q_{\phi_{\hat{\bx}}}(\bz|\hat{\bx})$ to $\bz$ instead of $p_{\phi_{\bx}}(\bz|{\bx})$ as shown currently in line 9 of Alg.~\ref{alg:alg_original}.  A full derivation of the dVAE and its variants is given in the supplementary.

\begin{algorithm}[b!]
	\caption{dVAE learning for additional $\bz_{\bu}$.}
	\label{alg:alg_semi}
	\begin{algorithmic}[1]
		\REQUIRE $\bx,\by_1,\lambda_{\bx},\lambda_{\by_1},\beta,T_1,T_2,T_3$
		\ENSURE $\phi_{\bx},\phi_{\by_1},\phi_{\bu},\theta_{\bx},\theta_{\by_1}$
		\STATE Initialize $\phi_{\bx},\phi_{\by_1},\phi_{\bu},\theta_{\bx},\theta_{\by_1}$
		
		\FOR{$t_1 = 1 ,\dots, T_1$ epochs}
		\STATE{Encode $\bx,\by_1$ to $q_{\phi_{\by_1}}(\bz_{\by_1}|\by_1),q_{\phi_{\bu}}(\bz_{\bu}|\bx)$}
		\STATE{Construct $\bz \gets [\bz_{\by_1},\bz_\bu]$, $[\mu,\sigma] \gets q_{\phi_{\bu}}(\bz_\bu|\bx)$}
		\STATE{Decode $\bz$ to $p_{\theta_\bx}(\bx|\bz),p_{\theta_{\by_1}}(\by_1|\bz)$}
		\STATE{Update $\phi_{\by_1},\phi_{\bu},\theta_{\by_1},\theta_{\bx}$}
		
		\FOR{$t_2 = 1 ,\dots, T_2$ epochs}
		\STATE{Encode $\by_1$ to $q_{\phi_{\by_1}}(\bz_{\by_1}|\by_1)$}
		\STATE{Construct $\bz_{noise} \gets \mathcal{N}(\mathbf{\mu},\mathbf{\sigma})$, $\mathbf{z} \gets [\bz_{\by_1},\bz_{noise}]$}
		\STATE{Decode $\bz$ to $p_{\theta_{\by_1}}(\by_1|\bz$)}
		\STATE{Update $\phi_{\by_1},\theta_{\by_1}$}
		\ENDFOR
		
		\ENDFOR

		\FOR{$t_3 = 1 ,\dots, T_3$ epochs}
		\STATE{Encode $\bx$ to $q_{\phi_{\bx}}(\bz|\bx)$}
		\STATE{Construct $[\bz_{\by_1},\bz_\bu] \gets \bz$}
		\STATE{Decode $\bz$ to $p_{\theta_\bx}(\bx|\bz),p_{\theta_{\by_1}}(\by_1|\bz)$}
		\STATE{Update $\phi_{\bx}$}
		\ENDFOR
		
	\end{algorithmic}
\end{algorithm}

\begin{figure*}[t]
	\centering	
	\includegraphics[width=\linewidth]{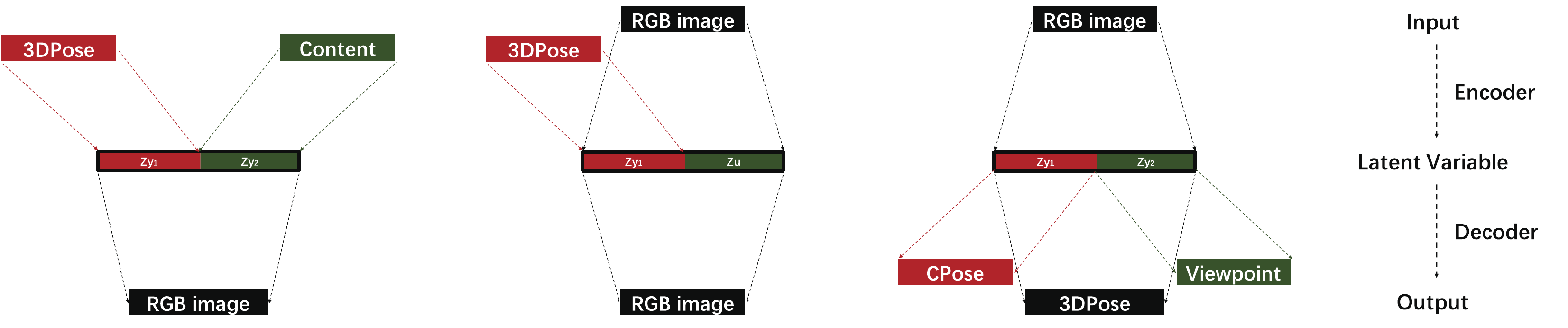}
	\caption{Inference models for the tasks of image synthesis (left and middle) and pose estimation (right). 
	}
	\label{fig:module}
\end{figure*}

\subsection{Applications}~\label{sec:applications}
Based on the theory proposed above, we develop two applications: image synthesis and pose estimation from RGB images. Like~\cite{zimmermann2017learning}, we distinguish between an absolute 3D hand pose (3DPose), a canonical hand pose (CPose), and a viewpoint.  The canonical pose is a normalized version of the 3D pose within the canonical frame, while viewpoint is the rotation matrix that rotates CPose to 3DPose. 


In \textbf{image synthesis}, we would like to sample values of $\bz$ and decode this into an image $\bx$ via the generative model $p_{\theta_{\bx}}$.  To control the images being sampled, we want to have a latent $\bz$ which is disentangled with respect to the 3DPose, and image (background) content, \ie all aspects of the RGB image not specifically related to the hand pose itself.  A schematic of the image synthesis is shown in the left panel of Fig.~\ref{fig:module}; in this case, we follow the model in Fig.~\ref{fig:dvae_simp} and use Alg.~\ref{alg:alg_original}.  Here, $\by_1$ would represent 3DPose and $\by_2$ would represent the image content; similar to~\cite{tulyakov2017mocogan}, this content is specified by a representative tag image.  
By changing the inputs $\by_1$ and $\by_2$, \ie by varying the 3DPose and content through the encoders $q_{\phi_{\by_1}}$ and $q_{\phi_{\by_2}}$, we synthesize new images with specified poses and background content.  Furthermore, we can also evaluate the pose error of the synthesized image via the pose decoder $p_{\theta_{\by_1}}$.



Tag images for specifying background content are easy to obtain if one has video sequences from which to extract RGB frames. However, for some scenarios, this may not be the case, \ie if each RGB image in the training set contains different background content.  This is what necessitates the model in Fig.~\ref{fig:dvae_gen} and the learning algorithm in Alg.~\ref{alg:alg_semi}. In such a scenario, $\by_1$ again represents the 3DPose, while the image content is modelled indirectly through $\bx$.  For testing purposes, however, there is no distinction between the two variants, as input is still given in the form of a desired 3DPose and an RGB image specifying the content. 


For \textbf{hand pose estimation}, we aim to predict 3DPose $\bx$, CPose $\by_1$ and viewpoint $\by_2$
from RGB image $\hat{\bx}$ according to the model in Fig.~\ref{fig:dvae_simp2} by disentangling $\bz$ into the CPose $\bz_{\by_1}$ and viewpoint $\bz_{\by_2}$. In this case, we embed $\bx$ and $\hat{\bx}$ into a shared latent space. We apply inference as shown by the right panel in Fig.~\ref{fig:module} and learn the model with Alg.~\ref{alg:alg_original}. Moreover, because annotated training data is sparse in real world applications, we can further leverage unlabelled or weakly labelled. 
Our proposed method consists of multiple VAEs, which can be trained respectively for semi- and weakly-supervised setting. For semi-supervised setting, we use both labelled and unlabelled CPose, viewpoint and 3DPose data to train the encoders $q_{\phi_{\by_1}}$,$q_{\phi_{\by_2}}$ and all decoders in the disentangled step. For weakly-supervised setting, we exploit images and their weak labels like viewpoint $\by_2$ by training the VAE with $q_{\phi_{\hat{\bx}}}$ and $p_{\theta_{\by_2}}$ in the embedding step.

\begin{figure*}[htbp]
	\includegraphics[width=\linewidth]{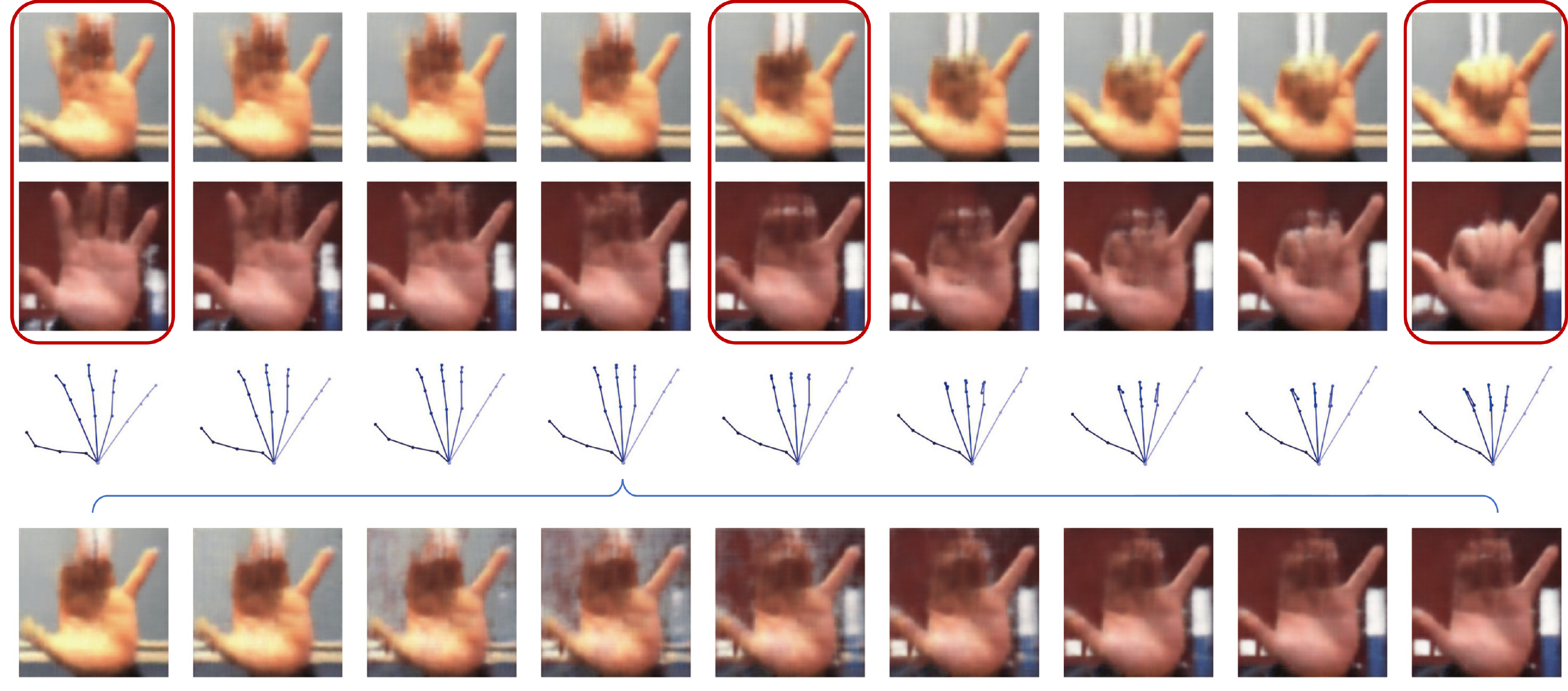}
		\caption{Latent space walk. The images in the red boxes are provided inputs. The first two rows show synthesized images when interpolating on the latent 3DPose space; the third row shows skeletons of the reconstructed 3DPose. The fourth row shows synthesized images when the pose is fixed (to the fourth column) when interpolating in the content latent space. 
		}
	\label{fig:randomwalk}
\end{figure*}

\section{Experimentation}
A good disentangled representation should show good performance on both discriminative tasks such as hand pose estimation as well as  generative tasks. We transfer attributes between images and infer 3D hand poses from monocular hand RGB images via disentangled representations. More precisely, for image synthesis, we transfer image content with fixed 3DPose, while for 3D hand pose estimation, we predict viewpoint, CPose  and 3DPose. 

\subsection{Implementation details}
Our architecture consists of multiple encoders and decoders. For encoding images, we use Resnet-18~\cite{he2016deep}; for decoding images, we follow the decoder architecture DCGAN~\cite{radford2015unsupervised}. 
For encoding and decoding hand poses, we use six fully connected layers with 512 hidden units. 
Exact architectural specifications are provided in the supplementary. 

For learning, we use the ADAM optimizer with a learning rate of $10^{-4}$, a batch size of 32.  We fix the dimensionality of $d$ of $\bz$ to 64 
and set the dimensionality of sub-latent variable $\bz_{\by_1}$ and $\bz_{\by_2}$ to 32 and 32.  For all applications, the $\lambda$'s are fixed ($\lambda_{\bx}\!=\!1, \lambda_{\by_1}\!=\!\lambda_{\by_2}\!=\!0.01$) while we must adjust $\beta$ ($\beta\!=\!100$ for image synthesis, $\beta'''\!=\!0.01$ for pose estimation). Further discussion on the impact of $\beta$ and $d$ can be found in the supplementary.



\subsection{Datasets \& Evaluation}
We evaluate our proposed method on two publicly available datasets: Stereo Hand Pose Tracking Benchmark (STB)~\cite{zhang20163d} and Rendered Hand Pose Dataset (RHD)~\cite{zimmermann2017learning}. 

The \textbf{STB dataset} features videos of a single person's left hand in front of 6 real-world indoor backgrounds. It provides the 3D positions of palm and finger joints for approximately 18k stereo pairs with 640 $\times$ 480 resolution. 
Image synthesis is relatively easy for this dataset due to the small number of backgrounds. To evaluate our model's pose estimation accuracy, we use the 15k / 3k  training/test split as given by ~\cite{zimmermann2017learning}. 
For evaluating our dVAE's generative modelling capabilities, we disentangle $\bz$ into two content and 3DPose according to the model in Fig.~\ref{fig:dvae_simp} synthesize images with fixed poses as per the left-most model in Fig.~\ref{fig:module}.

\textbf{RHD} is a synthesized dataset of rendered hand images with $320\!\times\!320$ resolution from 20 characters performing 39 actions with various hand sizes, viewpoints and backgrounds. The dataset is highly challenging due to the diverse visual scenery, illumination and noise. It is composed of 42k images for training and 2.7k images for testing. 

For quantitative evaluation and comparison with other works on 3D hand pose estimation, we use the common metrics, mean end-point-error (EPE) and the area under the curve (AUC) on the percentage of correct keypoints (PCK) score. Mean EPE is defined as the average euclidean distance between predicted and groundtruth keypoints; PCK is the percentage of predicted keypoints that fall within some given distance with respect to the ground truth.

\begin{figure*}[t!]
	\includegraphics[width=\linewidth]{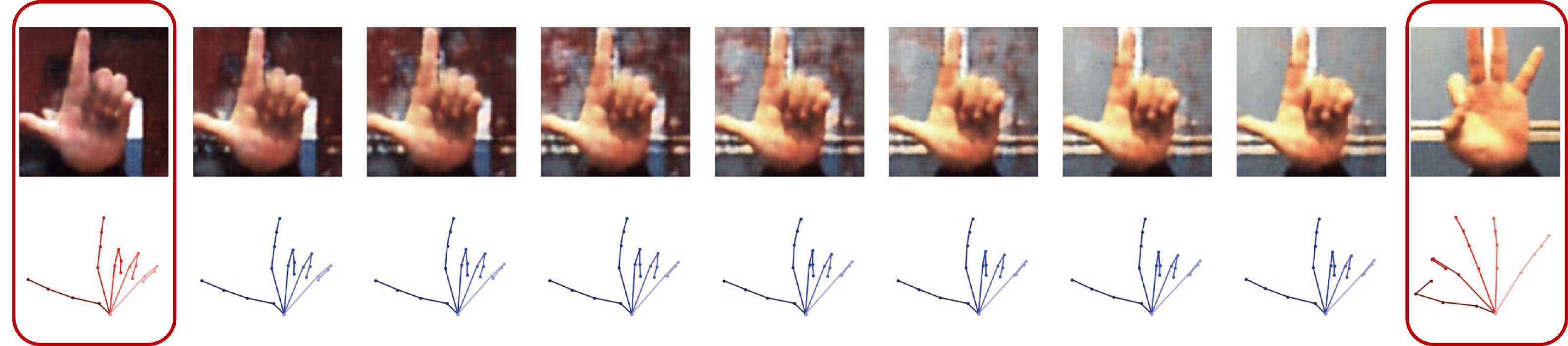}
	\caption{Latent space walk, interpolating $\bz_{\bu}$ representing image background content. The images along with groundtruth 3DPose (red) in the red box are the input points; the first row shows generated images and the second row corresponding reconstructed 3DPose (blue).  Note that because we are interpolating only on the background content, the pose stays well-fixed.}
	\label{fig:random_walk_semi}
\end{figure*}

\subsection{Synthesizing Images}
We evaluate the ability of our model to synthesize images by sampling from latent space walks and by transferring pose from one image to another.  

For the \textbf{fully specified latent $\bz$} model 
we show the synthesized images (see Fig.~\ref{fig:randomwalk}) when we interpolate the 3DPose while keeping the image content fixed (rows 1-3) and when we interpolate image content while keeping the pose fixed.  In both latent space walks, the reconstructed poses as well as the synthesized images demonstrate a smoothness and consistency of the latent space. 

\begin{figure}[b!]
	\centering	
	\includegraphics[width=\linewidth]{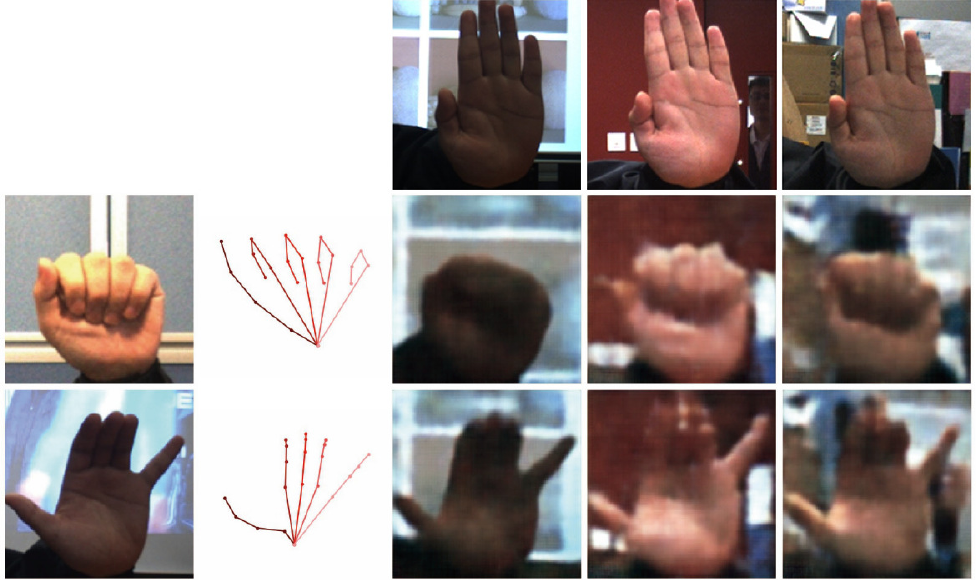}
	
	\caption{Pose transfer. The first column corresponds to images from which we extract the 3DPose (ground truth pose in second column); the first row corresponds to tag images columns we extract the latent content; the 2-3 rows, 3-5 columns are pose transferred images. 
	}
	\label{fig:posetransfer}
\end{figure}

We can also extract disentangled latent factors from different hand images and then recombine them to transfer poses from one image to another.  Fig.~\ref{fig:posetransfer} shows the results when we take poses from one image (leftmost column), content from other images (top row) and recombine them (rows 2-3, columns 3-5).  We are able to accurately transfer the hand poses while faithfully maintaining the tag content. 



\begin{figure*}[t]
		\centering
	\begin{minipage}[c]{0.3\textwidth} 
		\centering
		\includegraphics[width=\linewidth]{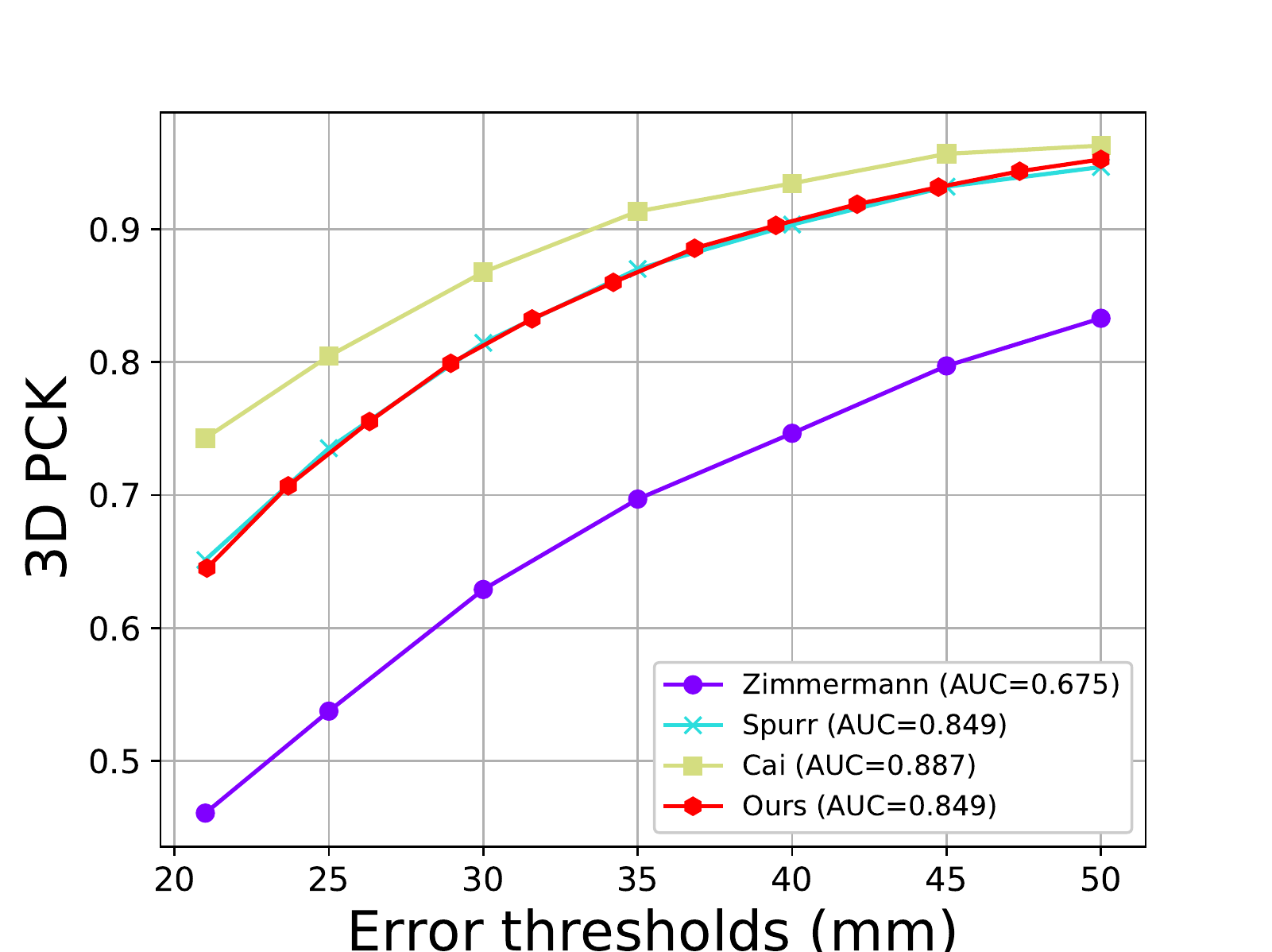}
	\end{minipage}
	\begin{minipage}[c]{0.3\textwidth} 
		\centering
		\includegraphics[width=\linewidth]{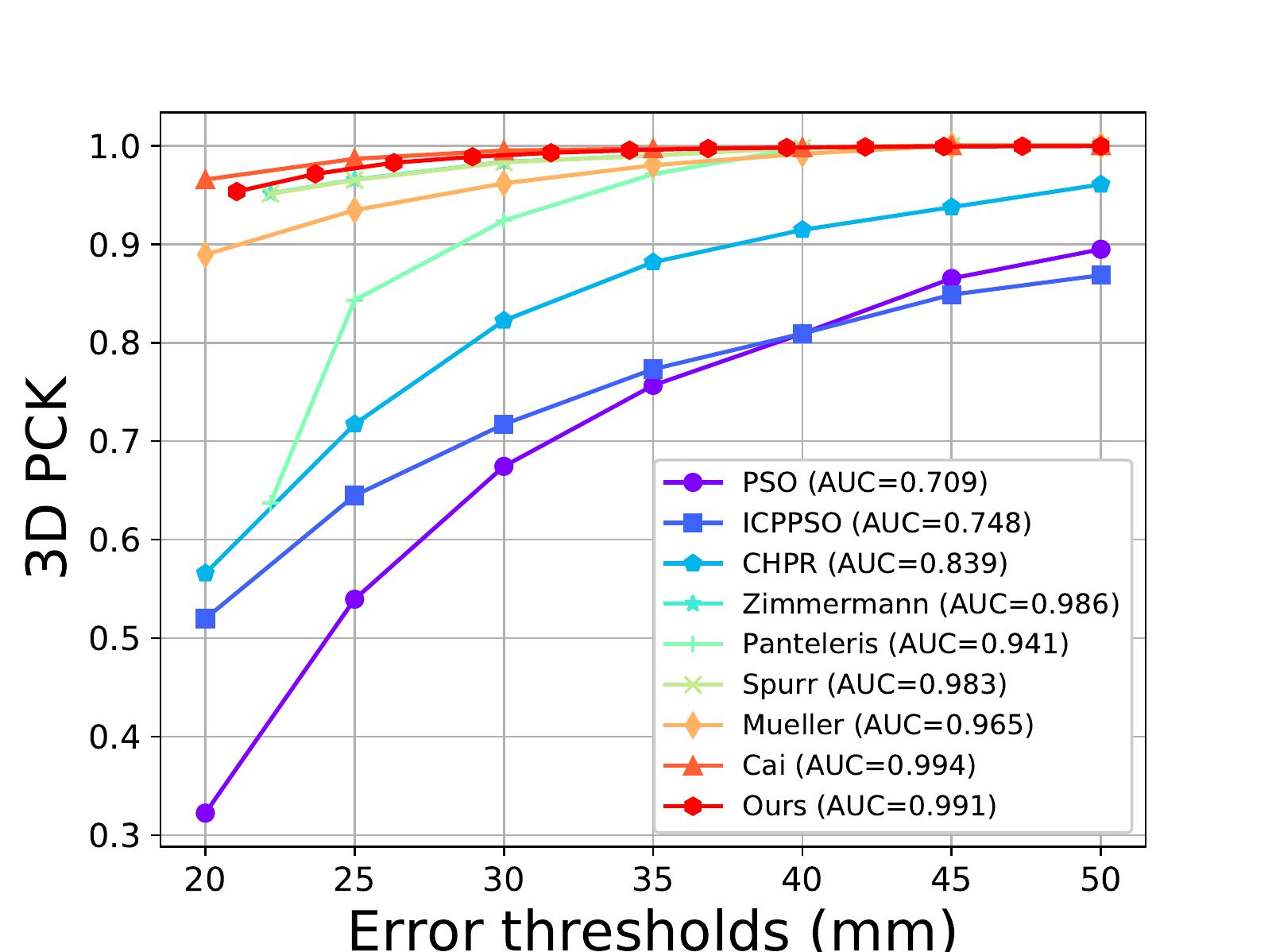}
	\end{minipage}
	\begin{minipage}[c]{0.35\textwidth} 
	\centering
	\scalebox{0.8}{
	\begin{threeparttable}
		\centering
		\begin{tabular}{ccccc}
			\toprule
			\multirow{2}{*}{Method}&
			\multicolumn{2}{c}{RHD}&\multicolumn{2}{c}{STB}\cr
			\cmidrule(lr){2-3} \cmidrule(lr){4-5}
			&CPose&3DPose&CPose&3DPose\cr
			\midrule
			\cite{zimmermann2017learning}&16.37&30.42&6.07&8.68\cr
			\cite{spurr2018cvpr}&$\backslash$&19.73&$\backslash$&8.56\cr
			Ours &13.93&19.95&6.09&8.66\cr
			\bottomrule
		\end{tabular}
	\end{threeparttable}}
    \end{minipage}	
    \hfill
	\caption{Quantitative evaluation. 3D PCK on RHD (left) and STB (middle). Mean EPE (mm) on RHD and STB (right).}
	\label{fig:evaluation}
	\end{figure*}

\textbf{With additional $\bz_{\bu}$} 
we also show interpolated results from a latent space walk on $\bz_{\bu}$ in Fig.~\ref{fig:random_walk_semi}.  In this case, the 3DPose stays well-fixed, while the content changes smoothly between the two input images, demonstrating our model's ability to disentangle the image background content even with out specific tag images for training.  

	\begin{figure*}[htb]
		\centering	
		\subfloat{	
			\centering
			 \includegraphics[width=\linewidth]{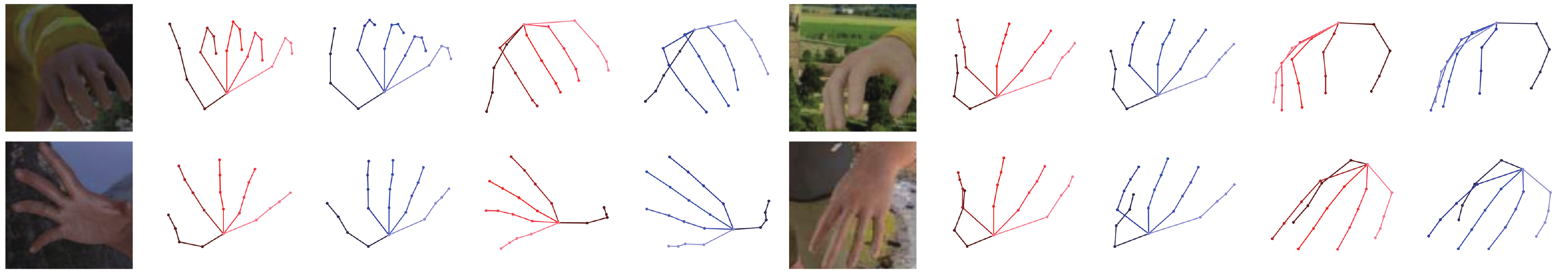}}

		\subfloat{
			\centering
			 \includegraphics[width=\linewidth]{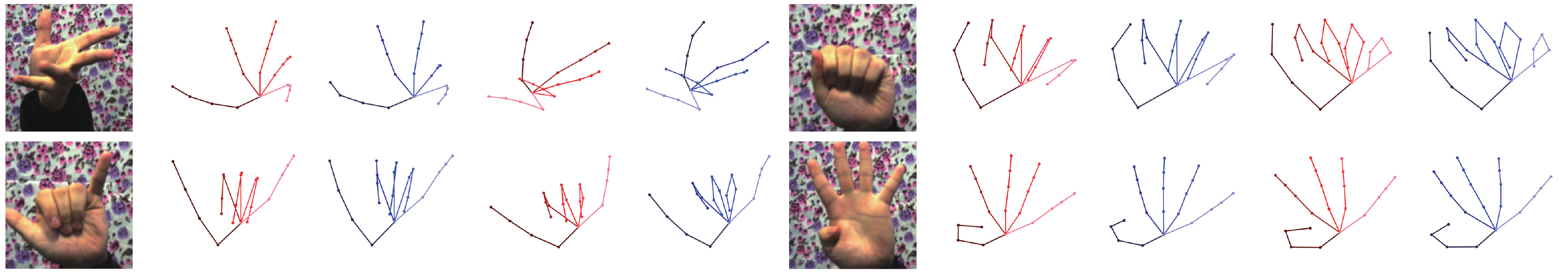}}
		\caption{CPose and 3DPose estimation on RHD and STB. For each quintet, the left most column corresponds to the input images, the second and the third columns correspond to CPose groundtruth (red) and our prediction (blue), the right most two columns correspond to 3DPose groundtruth (red) and our prediction (blue).}
		\label{fig:pose_result}
	\end{figure*}
	
\subsection{3D hand pose estimation}
We evaluate the ability of our dVAE to estimate 3D hand poses from RGB images based on the model variant described in Section~\ref{sec:applications} and compare against state-of-the-art methods~\cite{caiweakly,spurr2018cvpr,zimmermann2017learning,mueller2018ganerated,panteleris2017using} on both the RHD and STB datasets. In~\cite{zimmermann2017learning}, a two-stream architecture is applied to estimate viewpoint and CPose; these two are then combined to predict 3DPose.  
To be directly comparable, we disentangle the latent $\bz$ into a viewpoint factor and a CPose factor, as shown in Fig.~\ref{fig:module} right. Note that due to the decompositional nature of our latent space, we can predict viewpoint, CPose and 3DPose through one latent space. 

We follow the experimental setting in \cite{zimmermann2017learning,spurr2018cvpr} that left vs right handedness and scale are given at test time.  We augment the training data by rotating the images in the range of $[-180^{\circ},180^{\circ}]$ and making random flips along the $y$-axis while applying the same transformations to the ground truth labels. 
We compare the mean EPE 
in Fig.~\ref{fig:evaluation} right. 
We outperform~\cite{zimmermann2017learning} on both CPose and 3DPose. 
These results highlight the strong capabilities of our dVAE model for accurate hand pose estimation.
Our mean EPE is very close to that of~\cite{spurr2018cvpr}, while our 3D PCK is slightly better. As such, we conclude that the pose estimation capabilities of our model is comparable to that of~\cite{spurr2018cvpr}, though our model is able to obtain a disentangled representation and make full use of weak labels. 
We compare the PCK curves with state-of-the-art methods~\cite{caiweakly,spurr2018cvpr,zimmermann2017learning,mueller2018ganerated,panteleris2017using} on both datasets 
in Fig.\;\ref{fig:evaluation}. 
Our method is comparable or better than most existing methods except~\cite{caiweakly}, which has a higher AUC of 0.038 on RHD and 0.03 on STB for the PCK.  However, these results are not directly comparable, as~\cite{caiweakly} incorporate depth images as an additional source of training data. Fig.~\ref{fig:pose_result} shows some our estimated hand poses from both RHD and STB datasets. 

\textbf{Semi-, weakly-supervised learning:}
\label{sec:semi_supervised_dis_learning}
To evaluate our method in semi- and weakly-supervised settings, we sample the first $m\%$ images as labelled data and the rest as unlabelled data by discarding the labels of 3DPose, CPose and viewpoint.  We also consider using only viewpoints as a weak label while discarding 3DPose and CPose. For the RHD dataset, we vary $m\%$ from $5\%$ to $100\%$ and compare the mean EPE against the fully supervised setting. 
We can see that our model makes full use of additional information. With CPose, viewpoint and 3DPose labels, we improve the mean EPE up to $3.5\%$. With additional images and viewpoint labels, the improvement is up to $7.5\%$.

\begin{figure}[htb]
	\centering	
	\includegraphics[width=0.6\linewidth]{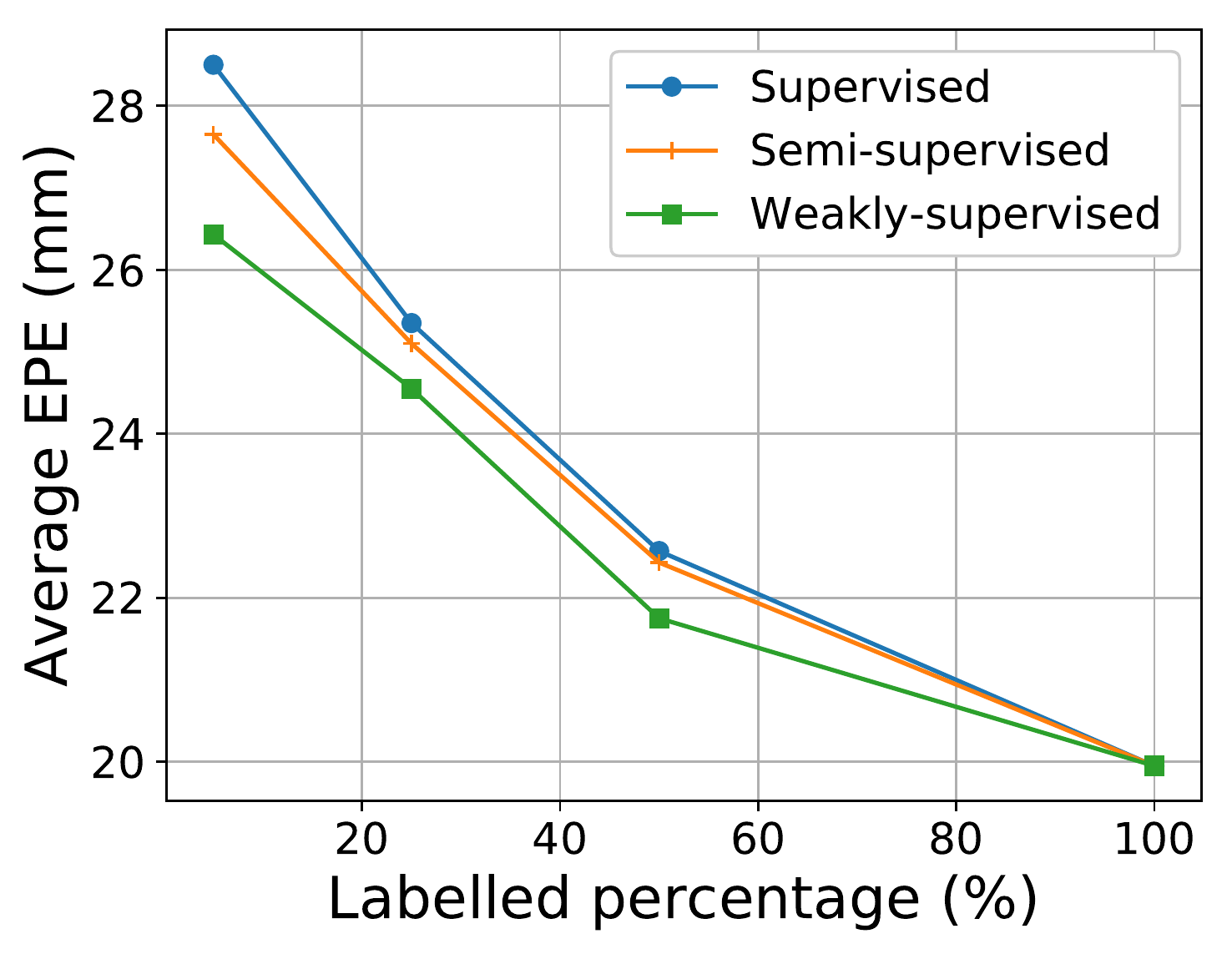}
	\caption{Mean EPE of our model on the semi-supervised setting and the weakly-supervised setting.}
	\label{fig:semi_EPE}
\end{figure}

\section{Conclusion}
In this paper, we presented a VAE-based method for learning disentangled representations of hand poses and hand images.  We find that our model allows us to synthesize highly realistic looking RGB images of hands with full control over factors of variation such as image background content and hand pose. However, the factors of variation here should be independent. This is a valid assumption for hand images, but we will consider to relax the need of independence between factors and further investigate disentangled representations with multimodal learning. 

For hand pose estimation, our model is competitive with state of the art and is also able to leverage unlabelled and weak labels.  Currently, STB is the standard benchmark for real-world monocular RGB hand pose estimation. However, since the featured background content and hand poses are quite simple, performance by state-of-the-art methods on this dataset has become saturated. For the 3D PCK, recent works~\cite{caiweakly,spurr2018cvpr,zimmermann2017learning,mueller2018ganerated,panteleris2017using} achieve AUC values for error thresholds of 20-50mm ranging from $96\%$ to more than $99\%$. As such, we encourage members of the community to collect more challenging benchmarks for RGB hand pose estimation.  In particular, for the monocular scenario, one possibility would be to collect multi-view~\cite{rhodin2018learning} and also multi-modal data, \ie RGBD, from which it is possible to use highly accurate model-based trackers to estimate ground truth labels.  

\vspace{-0.4cm}
\paragraph{Acknowledgments} Research in this paper was partly supported by the Singapore Ministry of Education Academic Research Fund Tier 1. We also gratefully acknowledge NVIDIA's donation of a Titan X Pascal GPU.



{\small
\bibliographystyle{ieee}
\bibliography{vaehand}
}

\end{document}